\pdfoutput=1

\documentclass[11pt]{article}

\usepackage[]{acl}

\usepackage{times}
\usepackage{mathtools}
\usepackage{latexsym}

\usepackage[T1]{fontenc}

\usepackage[utf8]{inputenc}

\usepackage{microtype}

\usepackage{CJKutf8}

\usepackage{graphicx}

\usepackage{tabularray}
\usepackage{tabularx}
\usepackage{multirow}
\usepackage{pifont}
\usepackage{makecell}
\usepackage{booktabs}

\usepackage{hyperref}

\usepackage{comment}

\title{Surveying the Dead Minds: Historical-Psychological Text Analysis with Contextualized Construct Representation (CCR) for Classical Chinese}

\author{Yuqi Chen \\
        Peking University \\
        \texttt{cyq0722@pku.edu.cn} \\\And
        Sixuan Li \\
        Xiaoying AI Lab \\
        \texttt{lisixuan@xiaoyingai.com} \\\AND
        Ying Li \\
        Peking University \\
        \texttt{yingliclaire@pku.edu.cn} \\\And
        Mohammad Atari \\
        University of Massachusetts Amherst \\
        \texttt{matari@umass.edu} \\}

\begin{document}
\maketitle
\begin{abstract}
In this work, we develop a pipeline for historical-psychological text analysis in classical Chinese. Humans have produced texts in various languages for thousands of years; however, most of the computational literature is focused on contemporary languages and corpora. The emerging field of historical psychology relies on computational techniques to extract aspects of psychology from historical corpora using new methods developed in natural language processing (NLP). The present pipeline, called Contextualized Construct Representations (CCR), combines expert knowledge in psychometrics (i.e., psychological surveys) with text representations generated via transformer-based language models to measure psychological constructs such as traditionalism, norm strength, and collectivism in classical Chinese corpora. Considering the scarcity of available data, we propose an indirect supervised contrastive learning approach and build the first Chinese historical psychology corpus (C-HI-PSY) to fine-tune pre-trained models. We evaluate the pipeline to demonstrate its superior performance compared with other approaches. The CCR method outperforms word-embedding-based approaches across all of our tasks and exceeds prompting with GPT-4 in most tasks. Finally, we benchmark the pipeline against objective, external data to further verify its validity.
\end{abstract}

\begin{figure}[htbp]
  \centering
  \includegraphics[width=\linewidth]{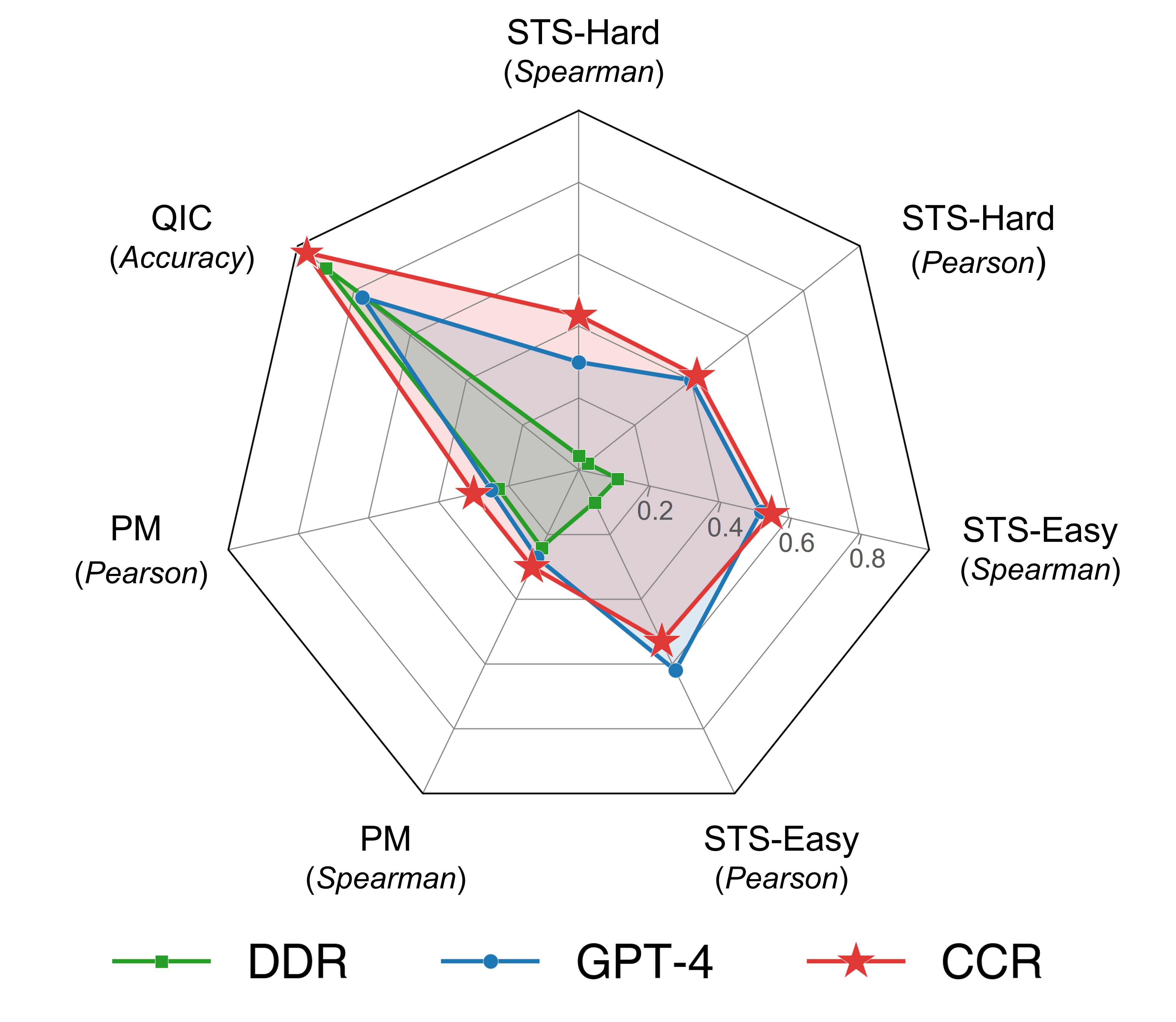}
  \caption{Comparison of the best performance among the DDR, CCR, and prompting methods on three tasks in the C-HI-PSY test set. (STS: Semantic Textual Similarity, PM: Psychological Measure, QIC: Questionnaire Item Classification)}
  \label{method_comparison}
\end{figure}

\begin{figure*}[!htp]
  \centering
  \includegraphics[width=\linewidth]{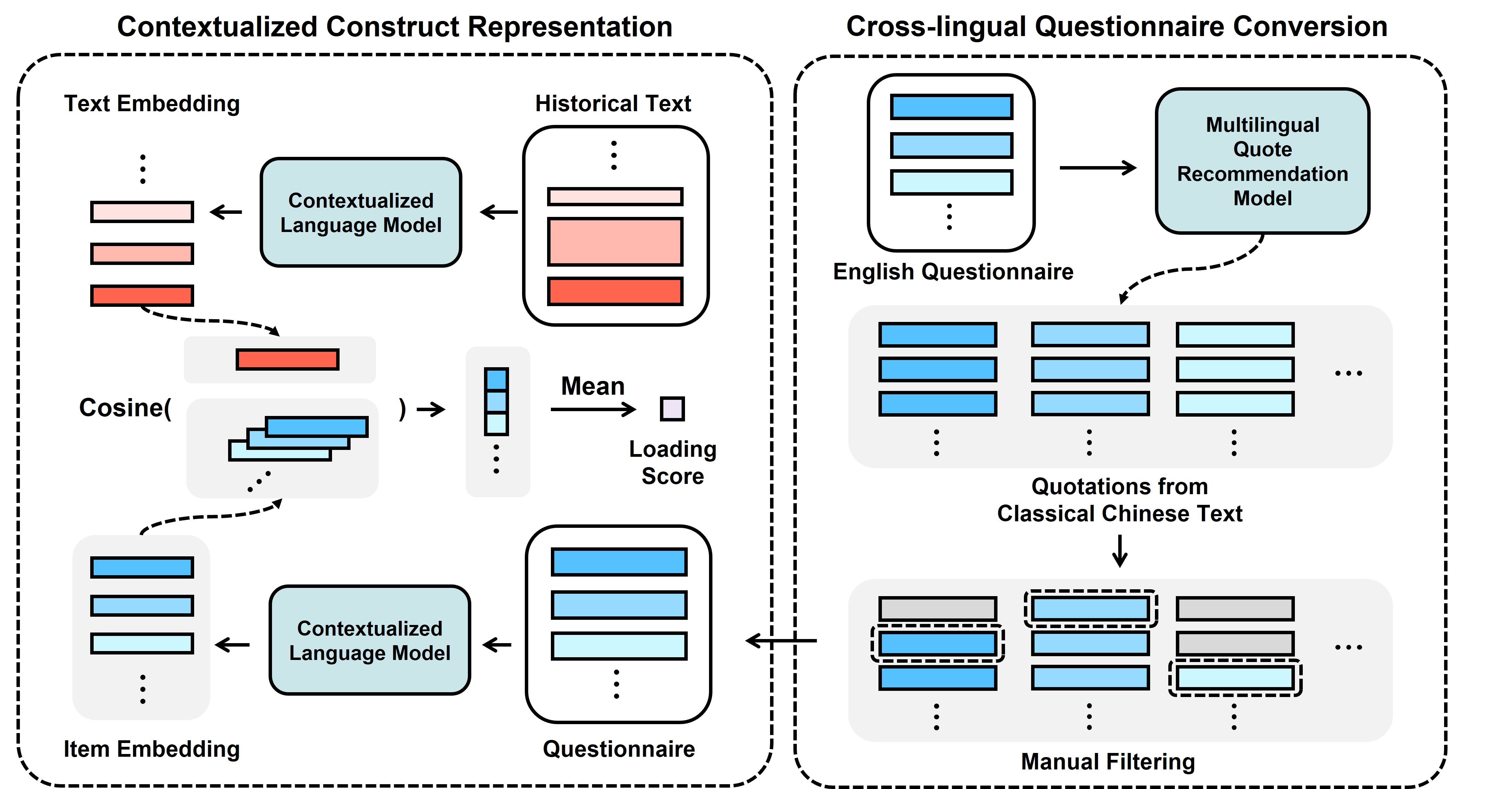}
  \caption{Pipeline of cross-lingual questionnaire conversion and contextualized construct representation for classical Chinese.}
  \label{ccr_for_classical_chinese}
\end{figure*}

\section{Introduction}

Humans have been producing written language for thousands of years. Historical populations have expressed their norms, values, stories, songs, and more in these texts. Such historical corpora represent a rich yet underexplored source of psychological data that contains the thoughts, feelings, and actions of people who lived in the past \citep{Jackson_2021}. The emerging field of ``historical psychology'' has been developed to understand how different aspects of psychology vary over historical time and how the origins of our contemporary psychology are rooted in historical processes \citep{Atari_2023,Muthukrishna_2021,Baumard_2024}. Since we cannot access ``dead minds'' directly but can access their textual remains, natural language processing (NLP) is the primary method to extract aspects of psychology from historical corpora. Previous works, however, are often monolingual and in English \citep{Blasi_2022}. In addition, much of the literature at the intersection of psychology and NLP has relied on bag-of-words or word embedding models, focusing on non-contextual word meanings rather than a holistic approach to language modeling.

Recently, more research attention in the NLP community has been directed to historical and ancient languages \citep{johnson-etal-2021-classical}, including but not limited to English \cite{manjavacas-arevalo-fonteyn-2021-macberth}, Latin \citep{Bamman2020LatinBA}, ancient Greek \citep{yousef-etal-2022-automatic-translation}, and ancient Hebrew \citep{swanson-tyers-2022-handling}. While all these languages have historical significance, classical Chinese is particularly important in the quantitative study of history. China has a long history spanning thousands of years, largely recorded in classical Chinese. The language served as a medium for expressing and disseminating influential philosophical and religious ideas. Confucianism, Daoism, and later Buddhism (through translations from Sanskrit) all found expression in classical Chinese, profoundly shaping Chinese thought, ethics, governance, and norms. As more resources become readily available for classical Chinese, scholars of ancient China can test more specific hypotheses using computational methods \citep{liu2023automatic,slingerland2013body,slingerland2017distant}.  

Due to its historical significance and geographical coverage, classical Chinese represents one of the most important languages in historical psychology \citep{Atari_2023}. Prior work in social science has often relied on bag-of-words approaches \citep{Zhong2023TheEO} or bottom-up techniques such as topic modeling \citep{slingerland2017distant}. In the NLP community, various Transformer-based models for classical Chinese have been developed \citep{tian2021anchibert, wang2022uncertainty, guwenbert, wang2023gujibert}, primarily for tasks like punctuation prediction \citep{zhou2023wyweb}, poem generation \citep{tian2021anchibert}, and translation \citep{wang-etal-2023-pre-trained}. However, they have not been applied to theory-driven psychological text analysis for extracting psychological constructs (e.g., moral values, norms, cultural orientation, mental health, religiosity, emotions, and thinking styles) from historical data. 

Transformer-based language models \citep{vaswani2017attention} are crucial for psychological text analysis because psychological constructs are often complex, and sentence-level semantics (and above) will more effectively capture psychological meanings than isolated words \citep{demszky2023using} or non-contextual word embedding models \citep{kennedy2021moral}. 

Here, we create a pipeline called Contextualized Construct Representation (CCR) for historical-psychological text analysis in classical Chinese. Although CCR has recently been developed for contemporary psychological text analysis \citep{atari2023ccr}, it can be adapted for historical NLP. As a tool for psychological text analysis, CCR takes advantage of contextual language models, does not require selecting a priori lists of words to represent a psychological construct (e.g., the popular Linguistic Inquiry and Word Count program, \citealp[]{boyd2022development}), and takes advantage of psychometrically validated questionnaires in psychology. 
The pipeline of CCR for classical Chinese proceeds in five steps: 
(1) selecting a questionnaire for the psychological construct of interest; 
(2) converting the questionnaire, usually in English, into classical Chinese;
(3) representing questionnaire items as embeddings using a contextual language model; 
(4) generating the embedding of the target text using a contextual language model; 
(5) computing the cosine similarity between the item and text embeddings. This straightforward pipeline is particularly useful for social science, wherein researchers are interested in interpretability and hypothesis testing.

There are two main challenges of using the CCR pipeline in analyzing Chinese historical texts: (1) popular self-report questionnaires, widely accepted by psychologists, are often in English, making it difficult to align them with classical Chinese texts; 
(2) there is a lack of psychology-specific Transformer-based models for classical Chinese, making it difficult to obtain high-quality representations of Chinese historical texts. To address the first challenge, we propose a pipeline that uses a multilingual quotation recommendation model \citep{qi2022quoter} to convert contemporary English questionnaires into contextually meaningful classical Chinese sentences (Section \ref{sec:cqc}). To tackle the second challenge, we build the first Chinese historical psychology corpus (C-HI-PSY) and introduce an approach based on indirect supervision \citep{he2021foreseeing, yin2023indirectly, xu2023nli} and contrastive learning \citep{chopra2005cl, schroff2015facenet, gao2021simcse, chuang2022diffcse} to fine-tune pre-trained models on this corpus (Section \ref{sec:iscl}).

\section{Related Work}

\paragraph{Psychological Text Analysis}
Given the increasing amount of online textual data, many social scientists are turning to NLP to test their theories. Unlike in some computational fields, social scientists traditionally give primacy to ``theory'' rather than prediction \citep{yarkoni2017choosing}. Hence, theory-driven text analysis is the first methodological choice in social sciences, including psychology \citep{Jackson_2021,wilkerson2017large,boyd2021natural}. Given the importance of theory development and hypothesis testing, many social scientists have developed dictionaries to assess psychological constructs as diverse as moral values \citep{graham2009liberals}, stereotypes \citep{nicolas2021comprehensive}, polarization \cite{simchon2022troll}, and threat \citep{choi2022danger}.   

\paragraph{Distributed Dictionary Representation (DDR)}

Aiming to integrate psychological theories with the capabilities of word embeddings, \citet{garten2018dictionaries} proposed the Distributed Dictionary Representation (DDR) as a top-down psychological text-analytic method. This method involves (a) defining a concise list of words by experts to capture a specific concept, (b) using a word-embedding model to represent these individual words, (c) computing the centroid of these word representations to define the dictionary's representation, (d) determining the centroid of the word embeddings within a given document, and (e) assessing the cosine similarity between the dictionary's representation and that of the document. DDR has been a useful approach in measuring moral rhetoric \cite{wang2021moral}, temporal trends in politics \citep{xu2023tracking}, and situational empathy \citep{zhou2021language}.

\paragraph{Contextualized Construct Representation (CCR)}

The Contextualized Construct Representation (CCR) pipeline is built upon SBERT \citep{reimers2019sentence}. This theory-driven and flexible approach has been shown to outperform dictionary-based methods and DDR for various psychological constructs such as religiosity, moral values, individualism, collectivism, and need for cognition \citep{atari2023ccr}. Furthermore, recent work suggests that CCR performs on par with Large Language Models (LLMs) such as GPT4 \citep{achiam2023gpt} in measuring psychological constructs \citep{abdurahman2023perils}. Although CCR has not been developed specifically for historical psychology, its flexible pipeline and easy-to-implement steps offer a unique opportunity to extract psychological constructs from historical corpora. In a way, CCR is similar to DDR, but instead of relying on non-contextual word embeddings, it makes use of the power of contextual language models to represent whole sentences (or larger texts). In addition, it obviates the development of researcher-curated word lists; instead, making use of thousands of existing questionnaires (which typically include face-valid declarative sentences with which participants agree or disagree) that have been developed and validated in psychology over the last century.

\paragraph{Semantic Textual Similarity}

While BERT \citep{Devlin2018BERTPO} can identify sentences with similar semantic meanings, this process can be resource-intensive. To enhance the performance of BERT for tasks like semantic similarity assessments, clustering, and semantic-based information retrieval, \citet{reimers2019sentence} developed Sentence-BERT (or SBERT). This model employs a Siamese network structure specifically designed to create embeddings at the sentence level. SBERT outperforms conventional transformer-based models in tasks related to sentences and significantly reduces the time needed for computations. It is engineered to generate sentence embeddings that capture the core semantic content, ensuring that sentences with comparable meanings are represented by closely positioned embeddings in the vector space. Therefore, SBERT provides an efficient and less computationally demanding method for evaluating semantic similarities between sentences, making it particularly useful in fields such as psychology \citep{juhng2023discourse,sen2022depression}.

\section{Methodology}

Employing the CCR pipeline for historical-psychological text analysis necessitates the use of valid questionnaires and appropriate contextual language models that can effectively represent sentences or paragraphs. We propose two distinct pipelines: (1) a cross-lingual questionnaire conversion pipeline to obtain psychological questionnaires in classical Chinese; (2) an indirect supervised contrastive learning pipeline to fine-tune pre-trained Transformer-based models using a historical psychological corpus.

\subsection{Cross-lingual Questionnaire Conversion}
\label{sec:cqc}

In order to calculate semantic similarities between questionnaires, typically in English, and the Chinese historical texts to be measured, typically in classical Chinese, we introduce a novel workflow for Cross-lingual Questionnaire Conversion (CQC).
Instead of relying on translations or generated texts, we employ quotations from authentic historical texts, as they can integrate more naturally within the context of classical Chinese.

The process of converting a contemporary English questionnaire $\mathcal{Q}$ into a classical Chinese questionnaire $\Tilde{\mathcal{Q}}$ is illustrated in the right panel of Figure \ref{ccr_for_classical_chinese}.
For each questionnaire item ($q_i \in \mathcal{Q}$), the multilingual quote recommendation model, ``QuoteR'' \citep{qi2022quoter}, which is trained on a dataset that includes English, modern Standard Chinese, and classical Chinese, can identify a set of quotations $\{\Tilde{q}\}_i$ in classical Chinese that are semantically similar to the English sentence $q_i$.

All the items are entered into the model for each questionnaire, resulting in a pool of corresponding quotations. Then, manual filtering is followed to eliminate quotations of low quality, which can be either inappropriate or not explicitly relevant to the psychological construct. Ultimately, the most similar quotations $\Tilde{q}_i$ are selected, substituting for every English $q_i$ to construct $\Tilde{\mathcal{Q}}$ in classical Chinese.

\subsection{Indirect Supervised Contrastive Learning}
\label{sec:iscl}

\begin{figure}[!htbp]
  \centering
  \includegraphics[width=\linewidth]{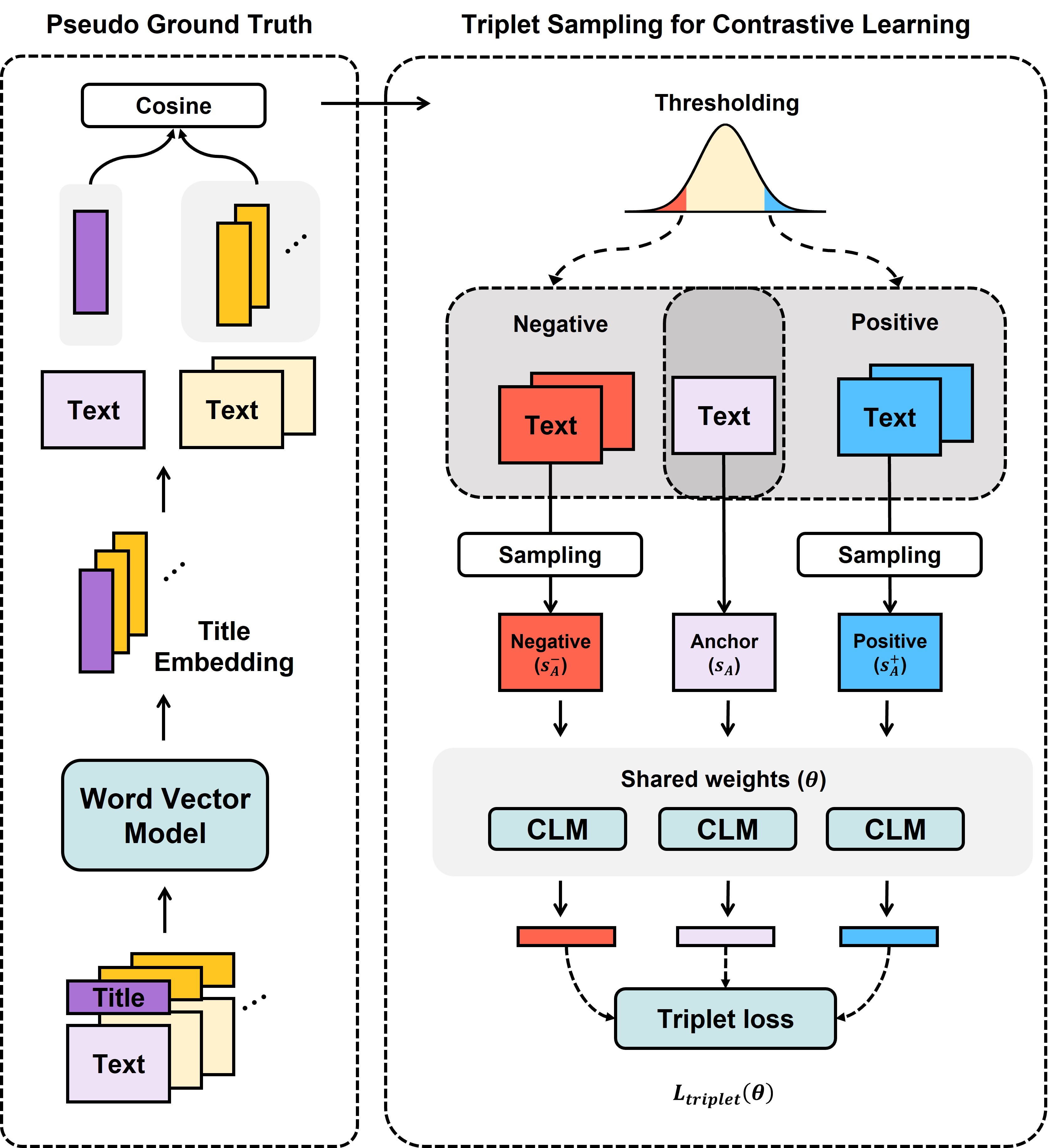}
  \caption{Pipeline of triplet sampling and contrastive learning. CLM stands for contextual language model.}
  \label{contrastive_learning}
\end{figure}

To obtain better psychology-specific representations for CCR in Chinese historical texts, we introduce an indirect supervised contrastive learning approach to finetune pre-trained Transformer-based models, as shown in Figure \ref{contrastive_learning}. 

\paragraph{Historical Psychology Corpus}

\begin{CJK*}{UTF8}{bsmi}
We assemble a refined corpus named Chinese Historical Psychology Corpus (C-HI-PSY), 
which is comprised of 21,539 paragraphs ($\mathcal{S}$) extracted from 667 distinct historical articles and book chapters in classical Chinese. 
The titles of these works ($\mathcal{T}$, $\vert {\mathcal{T}} \vert \ll \vert \mathcal{S} \vert$), each carefully selected for their relevance to moral values, serve as labels for their topics, including ``節義'' (moral integrity), ``孝弟'' (filial piety and fraternal duty), ``盡忠'' (utmost loyalty), ``廉恥'' (sense of shame), ``清介'' (pure and incorruptible), and ``愛己'' (love oneself), among others. 
\end{CJK*}

We divide our data into training, validation, and testing sets, allocating 60\%, 20\%, and 20\% of the data to each set, respectively. The distribution of paragraph lengths across different sets is consistent, as shown in Figure \ref{para_length_distribution} in Appendix \ref{sec:appendix_corpus_para}.

\paragraph{Pseudo Ground Truth from Titles}

Since the title ($t_i \in \mathcal{T}$) of a paragraph ($s_i \in \mathcal{S}$) is a concise summary of the moral values reflected in the paragraph, the semantic similarity between titles, $\textbf{sim} (t_i, t_j)$, can be considered as the pseudo ground truth for the semantic similarity between corresponding paragraphs, $\textbf{sim} (s_i, s_j)$. The semantic similarity between titles can be obtained by embedding the titles via $E_T(\cdot)$ and calculating their cosine similarity $\textbf{cos} (E_T(t_i), E_T(t_j))$. To perform word embedding on the titles, we trained five word vector models on a large classical Chinese corpus containing over a billion tokens using different frameworks and architectures, and picked the best-performing one (see Appendix \ref{sec:appendix_word_vector_model} for word vector model details).

\paragraph{Positive and Negative Sampling}

\begin{figure}[!htbp]
  \centering
  \includegraphics[width=0.9\linewidth]{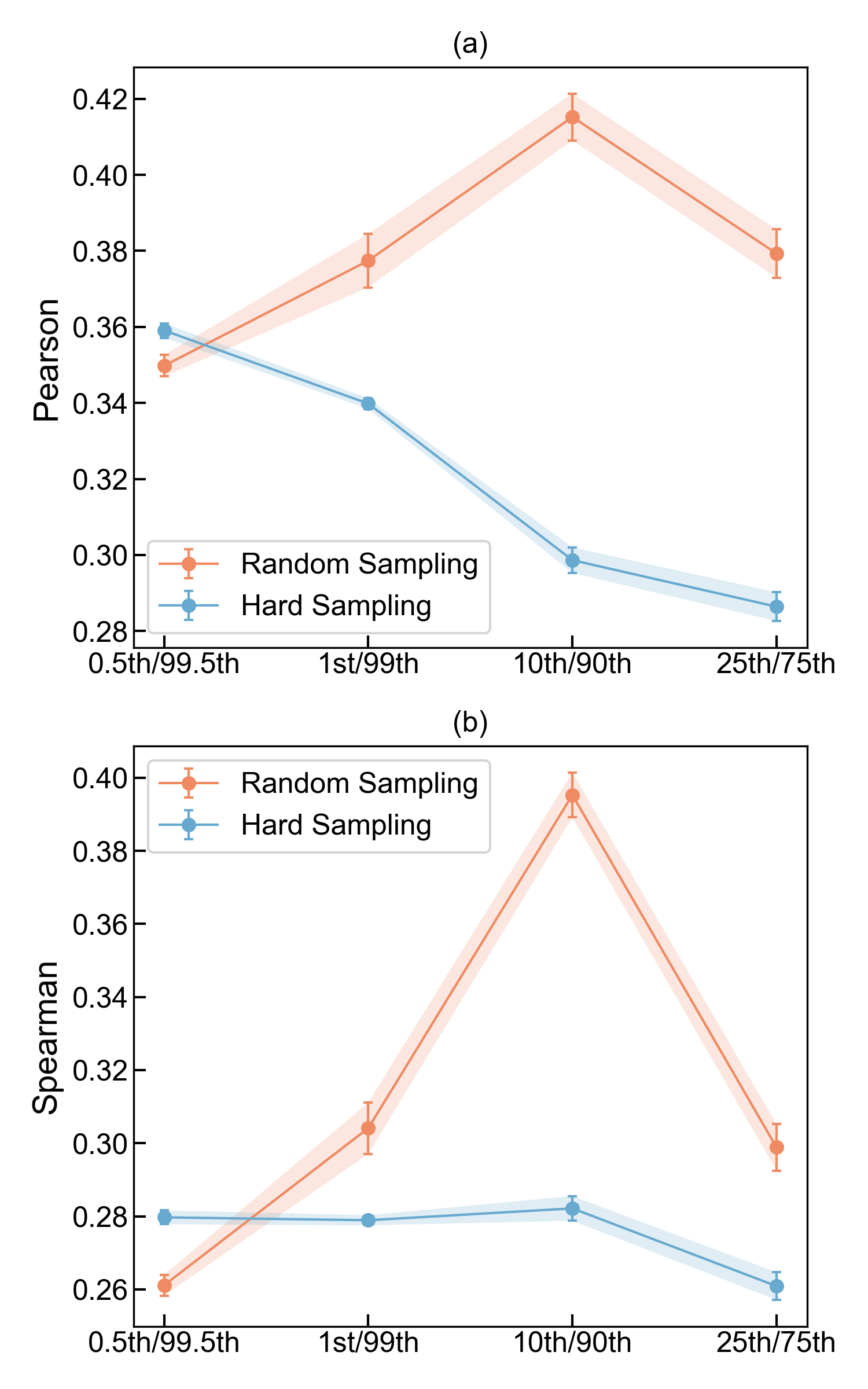}
  \caption{Performance variation with sampling methods and thresholds.}
  \label{sampling_compare}
\end{figure}

We calculate the cosine similarities between the title embeddings $\textbf{cos} (E_T(t_i), E_T(t_j))$, obtained through the word vector model, of all title pairs (the Cartesian product $\mathcal{T} \times \mathcal{T}$) in the corpus. The distribution of title similarities is illustrated in Figure \ref{title_similarity_distribution} in Appendix \ref{sec:appendix_corpus_title}. 
We obtain positive and negative paragraph pairs by thresholding the similarities of title pairs.
Paragraphs whose titles have similarities exceeding the upper threshold $\delta^+$, as well as those with identical titles, were identified as positive pairs $(\mathcal{S} \times \mathcal{S})^+$, that is,
$$
\{(s_i, s_j)^+\ |\ \textbf{sim} (E_T(t_i), E_T(t_j)) > \delta^+\} 
$$
Conversely, those with titles having similarities below the lower threshold $\delta^-$ were designated as negative pairs $(\mathcal{S} \times \mathcal{S})^-$, that is,
$$
\{(s_i, s_j)^-\  |\  \textbf{sim} (E_T(t_i), E_T(t_j)) < \delta^-\}
$$ 
We experiment with several threshold settings, including 0.5th/99.5th, 1st/99th, 10th/90th, and 25th/75th percentiles, on the C-HI-PSY validation set using the base model ``bert-ancient-chinese'' \citep{wang2022uncertainty}. Our findings demonstrate that the 10th/90th percentile threshold yields the best performance, see Figure \ref{sampling_compare}. Hence, for the following experiments, if not specified, the threshold setting has been taken as 10th/90th.

\paragraph{Triplet Sampling}

We implement two strategies, random sampling and hard sampling, to construct triplets of anchor-positive-negative paragraphs $(s_A^{}, s_A^+, s_A^-)$ from the training set.
In random sampling, we select one positive instance $s_A^+$ and one negative instance $s_A^-$ randomly from the respective positive pairs $(s_A \times \mathcal{S})^+$ and negative pairs $(s_A \times \mathcal{S})^-$ of the anchor $s_A$.
In hard sampling, we utilize the pre-trained model $f_{\theta}(\cdot)$, which is later fine-tuned on these triplets, to embed paragraphs and calculate cosine similarities between the positive and negative pairs as $\textbf{cos} (f_\theta (s_A^{}), f_\theta (s_A^{+/-}))$. For the positive instance, we choose the paragraph with the lowest similarity to the anchor from its positive pairs, that is,
\begin{equation*}
\begin{aligned}
    s_A^+ = \underset{s}{\mathrm{argmin}}\ 
    \{
    \textbf{cos} (f_\theta (s_A)&,\ f_\theta (s))\ |\\
    &(s_A, s) \in (s_A \times S)^+
    \}
\end{aligned}
\end{equation*}
Conversely, for the negative instance, we select the paragraph with the highest similarity to the anchor from its negative pairs, that is,
\begin{equation*}
\begin{aligned}
    s_A^- = \underset{s}{\mathrm{argmax}}\ 
    \{
    \textbf{cos} (f_\theta (s_A)&,\ f_\theta (s))\ |\\
    &(s_A, s) \in (s_A \times S)^-
    \}
\end{aligned}
\end{equation*}
To prevent the model from over-fitting, we ensure that each paragraph is used as an anchor only once, applying this rule across both random and hard sampling strategies. We also compare the two sampling procedures in Figure \ref{sampling_compare} with respect to each positive-negative splitting threshold. Interestingly, we find that the random sampling procedure is better than hard sampling ever since the threshold is higher/lower than 0.5th/99.5th; we note that the case could be due to the noise inevitably caused by the indirect supervised learning approach, which drove the hard sampling procedure to fail at finding helpful instances (see Limitation).

\paragraph{Fine-tuning with Contrastive Learning}

We fine-tune several pre-trained Transformer-based models \citep{wang2022uncertainty, guwenbert, reimers2019sentence, text2vec} on the C-HI-PSY training set, using a triplet loss function \citep{schroff2015facenet},

\begin{equation*}
\begin{aligned}
    L_{triplet}(\theta) = \sum_{s_A^{}\in S}{
        \textbf{max} \{\mathcal{D}^+ - \mathcal{D}^-, 0 \}
    }
\end{aligned}
\end{equation*}

where $\mathcal{D}^+$ denotes the distance between the positive pair, i.e. ${\Vert f_\theta (s_A^{}) - f_\theta (s_A^+) \Vert}_2^2$, and $\mathcal{D}^-$ denotes the distance between the negative pair, i.e. ${\Vert f_\theta (s_A^{}) - f_\theta (s_A^-) \Vert}_2^2$, $\alpha$ is a constant set to be 5, and $\theta$ stands for the pre-trained weights to be fine-tuned. This loss function aims to minimize the squared Euclidean norm between the anchor and positive, and maximize the squared Euclidean norm between the anchor and negative.

We construct paragraph pairs from the C-HI-PSY validation set through random sampling to validate the models during training, using the similarities between titles as pseudo ground truth to gauge the similarities between paragraphs. We perform a hyperparameter sweep (see Table \ref{hyperparameter_sweep} in the Appendix), to select the best-performing configuration for each model, as shown in Table \ref{best_perform_models}.

\begin{figure}[htbp]
  \centering
  \includegraphics[width=\linewidth]{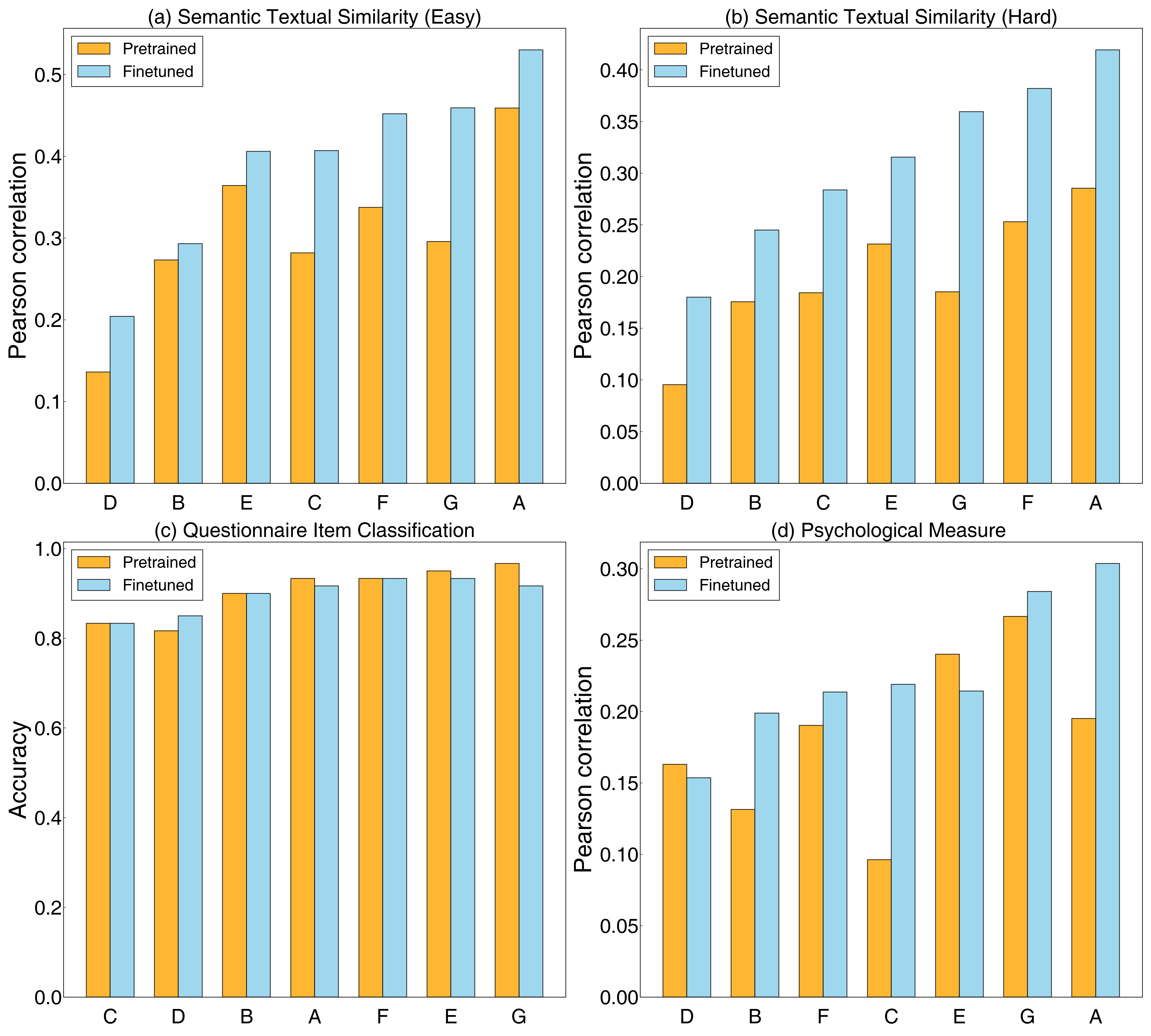}
  \caption{Comparison of model performance using the CCR method on the three tasks in the C-HI-PSY test set before and after fine-tuning. (Model A: bert-ancient-chinese, B: guwenbert-base, C: guwenbert-large, D: paraphrase-multilingual-MiniLM-L12-v2, E: text2vec-base-chinese, F: text2vec-base-chinese-paraphrase, G: text2vec-large-chinese)}
  \label{model_performance}
\end{figure}

\begin{table*}[!htbp]
\centering
\tabcolsep=0.2cm  %
\renewcommand\arraystretch{1.5}  %
\resizebox*{\linewidth}{!}{  %
\begin{tabular}{cccccccc}
\toprule[1.5pt]  %
Framework & Base Model & \makecell*[c]{If Specific to\\Classical Chinese} & \makecell*[c]{Batch\\Size} & \makecell*[c]{Warmup\\Epochs} & \makecell*[c]{Learning\\Rate} & Pearson & Spearman \\ \hline  %
BERT & Bert-ancient-chinese & \ding{52} & 32 & 3 & 1.0e-05 & \textbf{.43} & \textbf{.42} \\ \hline  %
\multirow{2}{*}{RoBERTa} & Guwenbert-base & \ding{52} & 32 & 2 & 2.0e-05 & .30 & .37 \\ \cline{2-8}
 & Guwenbert-large & \ding{52} & 16 & 1 & 2.0e-05 & .29 & .30 \\ \hline
SBERT & \makecell*[c]{Paraphrase-multilingual- \\ MiniLM-L12-v2} & \ding{55} & 32 & 1 & 2.0e-05 & .19 & .19  \\ \hline
MacBERT+CoSENT & text2vec-base-chinese & \ding{55} & 32 & 2 & 2.0e-05 & .34 & .32  \\ \hline
ERNIE+CoSENT & \makecell*[c]{text2vec-base-chinese-\\paraphrase} & \ding{55} & 32 & 2 & 2.0e-05 & .40 & .40  \\ \hline
LERT+CoSENT & text2vec-large-chinese & \ding{55} & 16 & 2 & 2.0e-05 & .36 & .37  \\
\bottomrule[1.5pt]  %
\end{tabular}
}
\caption{Fine-tuned models' performance on the validation set. We show the best performing configuration which is also the final configuration used to report each models' performance on the test test.}
\label{best_perform_models}
\end{table*}

\section{Evaluation and Results}
\label{sec: results}

We set up three different tasks to evaluate the CCR method (using SBERT models), and compare it with the DDR method (using word embedding models) and the prompting method (using generative LLMs).
The results are shown in Table \ref{psycho_measure}.

\subsection{Semantic Understanding}

\paragraph{Understanding of Historical Text: Semantic Textual Similarity}

For the CCR method, we embed whole paragraphs with SBERT models, and then calculate the cosine similarity between each pair of paragraphs. For the DDR method, we average the word vectors of all the words in the paragraph, and then calculate the cosine similarity between each pair of paragraphs. For the LLM-prompting method, we craft a few-shot prompt \citep{brown2020fewshot, si2022prompting} (Figure \ref{prompt_similarity}) asking for a similarity score, ranging from 0 to 1, between each pair of paragraphs. As mentioned, similarities between the titles of each pair of paragraphs are used as the pseudo ground truth.

We construct paragraph pairs for evaluation from the C-HI-PSY test set using two sampling methods: (1) random sampling, where paragraphs are randomly paired, and (2) threshold sampling, which pairs paragraphs with either positive or negative samples based on a specific threshold (10th/90th). Threshold sampling produces distinctly positive or negative pairs; thus, we refer to it as the Easy Task. Conversely, random sampling can result in ambiguous pairs, making for a more challenging Hard Task.

\paragraph{Understanding of Questionnaire Item: Text Classification}

We convert several broadly accepted questionnaires from English into classical Chinese, including Collectivism, Individualism \citep{Oyserman1993}, Norm Tightness and Norm Looseness \citep{Gelfand2011}, by employing the CQC approach described in Section \ref{sec:cqc}.
For both the CCR and DDR methods, all the items from these questionnaires are embedded. Then we conduct 10-fold cross-validation, using Support Vector Machines (SVM) as the classifier, and text embeddings or averaged word vectors as features. 
For the prompting method, we craft a few-shot prompt (Figure \ref{prompt_classification}) directly asking for classification.

\subsection{Psychological Measure}

For the CCR method, we calculate the average cosine similarities between each paragraph in the C-HI-PSY test set and all the items in the questionnaire, representing the ``loading score'' of the paragraph on the questionnaire.
For the DDR method, we build a corresponding dictionary for each psychological construct (see Appendix \ref{sec:appendix_dictionary} for more details), and calculate the cosine similarity between the centroid of words in each paragraph and the centroid of words in the dictionary.
For the prompting method, we craft a few-shot prompt (Figure \ref{prompt_measure}) asking for a score, ranging from 0 to 1, to measure each paragraph with respect to the topic of each questionnaire. Items in each questionnaire are provided in the prompt.
Average similarities between the title of each paragraph and all the words in the dictionary, calculated by the word vector model, are used as the pseudo ground truth.

\subsection{Results}

For the Semantic Textual Similarity (STS) task, we evaluate the DDR and CCR methods through a rigorous process involving 20 rounds of random sampling. In each round, 4,308 random paragraph pairs are constructed from the C-HI-PSY test set. After completing these 20 evaluations, we calculate the average scores along with standard errors. When evaluating the prompting method, due to the high costs, we only conduct a single round of random sampling. 
For the Questionnaire Item Classification (QIC) task, we utilize 60 items from questionnaires on Collectivism, Individualism \citep{Oyserman1993}, Norm Tightness, and Norm Looseness \citep{Gelfand2011}, selecting 15 items from each questionnaire. 
For the Psychological Measure (PM) task, we measure the loading scores of all 4308 paragraphs in the C-HI-PSY test set across the four questionnaires mentioned above, and report the average scores along with standard errors.

Figure \ref{model_performance} illustrates that the performance metrics of most models
in the CCR baseline have substantially improved after fine-tuning.
As shown in Table \ref{psycho_measure}, the CCR method, using SBERT models after fine-tuning, outperforms the DDR method across all tasks and surpasses the prompting method with GPT-4 (version January 25, 2024) in most tasks, demonstrating its superiority in effectively extracting psychological variables from text.

\begin{table*}[!ht]
\centering
\tabcolsep=0.1cm  %
\renewcommand\arraystretch{1.1}  %
\resizebox*{\linewidth}{!}{  %
\begin{tabular}{cccccccccc}
\toprule[1.5pt]  %
\multirow{2}{*}{Framework} & \multirow{2}{*}{Base Model} & \multicolumn{2}{c}{\makecell*[c]{Semantic\\Textual Similarity\\\small{(\textit{Easy Task})}}} & \multicolumn{2}{c}{\makecell*[c]{Semantic\\Textual Similarity\\\small{(\textit{Hard Task})}}} & \multicolumn{2}{c}{\makecell*[c]{Questionnaire\\Item Classification}} & \multicolumn{2}{c}{\makecell*[c]{Psychological\\Measure}} \\ \cmidrule(r){3-4} \cmidrule(r){5-6} \cmidrule(r){7-8} \cmidrule(r){9-10}
 &  & Pears. & Spear. & Pears. & Spear. & \multicolumn{2}{c}{Accuracy} & Pears. & Spear. \\  %
\hline\hline
\makecell*[c]{\textbf{\textit{(a) DDR}}\\ } \\
Word2Vec (CBOW) & / & $.02_{\pm .11}$ & $.02_{\pm .10}$ & $-.03_{\pm .02}$ & $-.02_{\pm .01}$ & \multicolumn{2}{c}{$.80_{\pm .16}$} & $.22_{\pm .07}$ & $.23_{\pm .05}$ \\
Word2Vec (Skip-gram) & / & $.08_{\pm .11}$ & $.09_{\pm .11}$ & $.02_{\pm .02}$ & $.02_{\pm .01}$ & \multicolumn{2}{c}{$.87_{\pm .15}$} & $.18_{\pm .07}$ & $.18_{\pm .06}$ \\
FastText (CBOW) & / & $.05_{\pm .11}$ & $.04_{\pm .10}$ & $-.01_{\pm .01}$ & $.01_{\pm .01}$ & \multicolumn{2}{c}{$.90_{\pm .13}$} & $.23_{\pm .08}$ & $.24_{\pm .06}$ \\
FastText (Skip-gram) & / & $.10_{\pm .10}$ & $.11_{\pm .10}$ & $.03_{\pm .02}$ & $.04_{\pm .01}$ & \multicolumn{2}{c}{$.85_{\pm .16}$} & $.20_{\pm .07}$ & $.20_{\pm .05}$ \\
\specialrule{0em}{0.1pt}{1.5pt} GloVe  & / & $.07_{\pm .10}$ & $.09_{\pm .11}$ & $.01_{\pm .02}$ & $.01_{\pm .01}$ & \multicolumn{2}{c}{$.83_{\pm .15}$} & $.16_{\pm .09}$ & $.19_{\pm .05}$ \\
\hline\hline
\makecell*[c]{\textbf{\textit{(b) Prompting}}\\ } \\
GPT & GPT-3.5-\textit{turbo-0125} & $.08$ & $.04$ & $.26$ & $.28$ & \multicolumn{2}{c}{$.63$} & $.05_{\pm .08}$ & $.08_{\pm .10}$ \\
GPT & GPT-4-\textit{0125-preview} & $\textbf{.62}$ & $.52$ & $.40$ & $.30$ & \multicolumn{2}{c}{$.77$} & $.25_{\pm .15}$ & $.27_{\pm .17}$ \\
\hline\hline
\makecell*[c]{\textbf{\textit{(c) CCR} (ours)}\\ } \\
BERT & Bert-ancient-chinese &  $.53_{\pm .07}$ & $\textbf{.55}_{\pm .07}$ & $\textbf{.42}_{\pm .01}$ & $\textbf{.43}_{\pm .01}$ & \multicolumn{2}{c}{$.93_{\pm .11}$} & $\textbf{.30}_{\pm .04}$ & $\textbf{.30}_{\pm .04}$ \\
RoBERTa & Guwenbert-base & $.29_{\pm .07}$ & $.46_{\pm .09}$ & $.25_{\pm .01}$ & $.40_{\pm .01}$ & \multicolumn{2}{c}{$.90_{\pm .11}$} & $.20_{\pm .06}$ & $.23_{\pm .09}$ \\
RoBERTa & Guwenbert-large & $.41_{\pm .05}$ & $.44_{\pm .07}$ & $.28_{\pm .01}$ & $.31_{\pm .01}$ & \multicolumn{2}{c}{$.83_{\pm .13}$} & $.22_{\pm .04}$ & $.20_{\pm .05}$ \\
SBERT & \makecell*[c]{Paraphrase-multilingual- \\ MiniLM-L12-v2} & $.20_{\pm .15}$ & $.21_{\pm .14}$ & $.18_{\pm .01}$ & $.19_{\pm .01}$ & \multicolumn{2}{c}{$.82_{\pm .19}$} & $.15_{\pm .04}$ & $.14_{\pm .05}$ \\
MacBERT+CoSENT & text2vec-base-chinese & $.41_{\pm .09}$ & $.40_{\pm .09}$ & $.32_{\pm .01}$ & $.31_{\pm .01}$ & \multicolumn{2}{c}{$.95_{\pm .08}$} & $.21_{\pm .10}$ & $.20_{\pm .10}$ \\ 
ERNIE+CoSENT & \makecell*[c]{text2vec-base-chinese-\\paraphrase} & $.45_{\pm .09}$ & $.45_{\pm .09}$  & $.38_{\pm .01}$ & $.37_{\pm .01}$ & \multicolumn{2}{c}{$.93_{\pm .11}$} & $.21_{\pm .03}$ & $.20_{\pm .04}$ \\ 
\specialrule{0em}{0.1pt}{1.5pt} LERT+CoSENT & text2vec-large-chinese & $.46_{\pm .12}$ & $.47_{\pm .08}$ & $.36_{\pm .01}$ & $.38_{\pm .01}$ & \multicolumn{2}{c}{$\textbf{.97}_{\pm .07}$} & $.28_{\pm .05}$ & $.27_{\pm .05}$ \\ 
\bottomrule[1.5pt]  %
\end{tabular}
}
\caption{Performance on the test set across three tasks using three methods: DDR, LLM Promping, and CCR. 
Details of models for the DDR method are explained in the Appendix \ref{sec:appendix_word_vector_model}.
Models for the CCR method have been fine-tuned on the C-HIS-PSY training set. Models for the prompting method include the versions of GPT-3.5 and GPT-4 that were released on January 25, 2024.}
\label{psycho_measure}
\end{table*}

\section{Benchmarking: Traditionalism, Authority, and Attitude toward Reform}
To address the lack of benchmark datasets related to psychological measurement in classical Chinese, we further validate the effectiveness of the CCR method using externally annotated data.

\paragraph{Officials' Attitudes toward Reform in the 11th Century}
Moral values and political orientations are closely intertwined \citep{federico2013mapping,kivikangas2021moral}.
For example, the attitude of individuals toward reforms, policy changes, and new legislation often reflects traditionalism, conservatism, and respect for authority \citep{hackenburg2023mapping,koleva2012tracing}. Those with stronger traditionalist views are more likely to identify with the existing social order and resist changes to the status quo \citep{osborne2023authoritarianism,jost2005antecedents}.

Throughout Chinese history, there have been numerous instances of significant reforms, one of the most notable of which being the Wang Anshi's New Policies in the 11th century, which faced mixed reactions from officials.
We draw upon a dataset manually compiled by \citet{wang2022blood}, who annotated the attitudes of 137 major officials toward the reform.

\paragraph{Individual-level Measure of Traditionalism and Authority}

We extract writings of these officials documented in the \textit{Complete Prose of the Song Dynasty} \citep{quansongwen2006}.
Questionnaires of traditionalism \citep{samore2023traditionalism} and authority \citep{atari2023mfq} are converted from English into classical Chinese, by employing the CQC approach described in Section \ref{sec:cqc}.
Employing the best-performing fine-tuned SBET model, we use our CCR pipeline to measure the levels of traditionalism and attitudes toward authority expressed in their texts. 
For each individual official, results are aggregated by calculating the average score across all of their writings.

\paragraph{Results}

\begin{figure}[htbp]
  \centering
  \includegraphics[width=\linewidth]{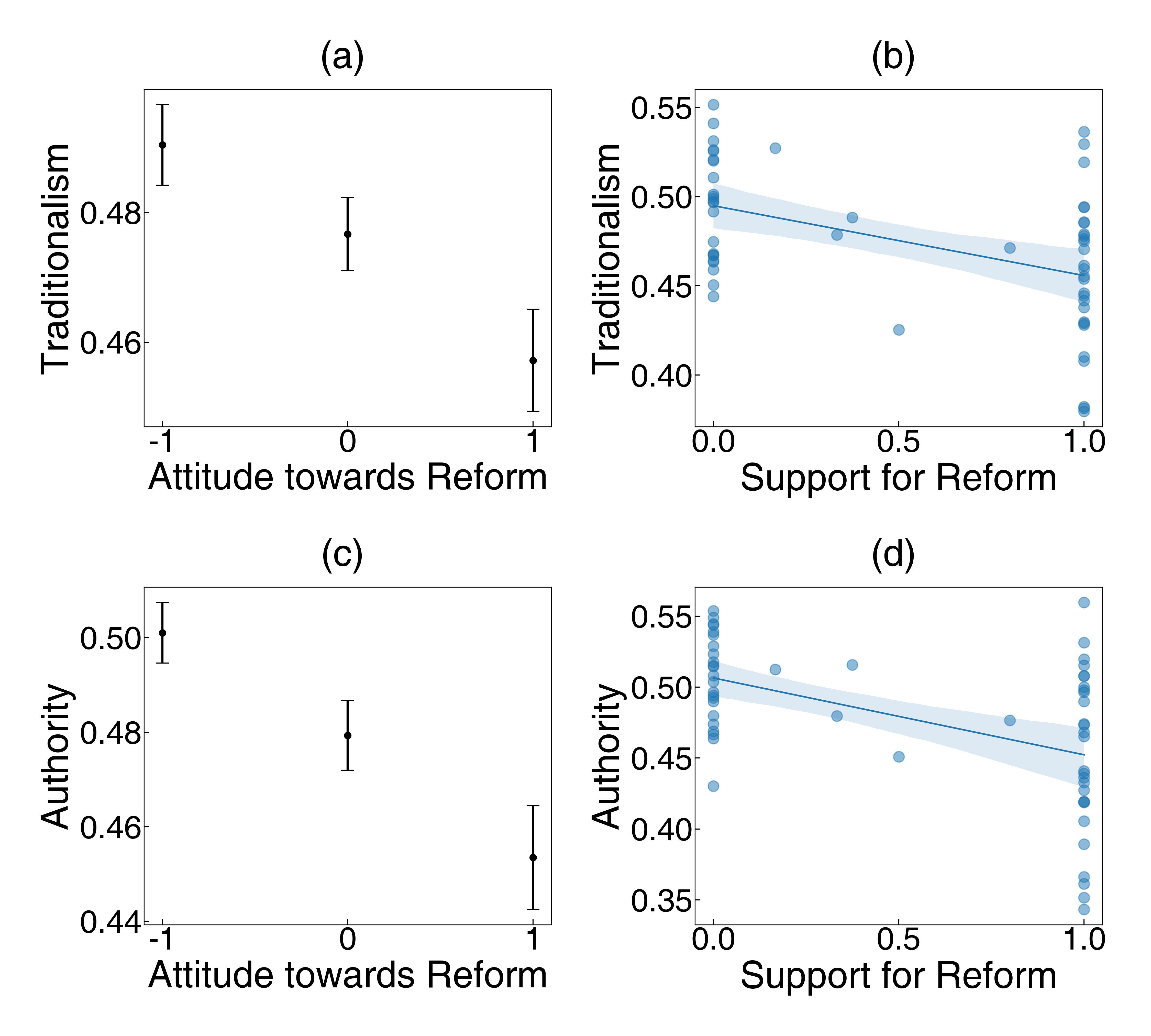}
  \caption{Correlation between Traditionalism  Authority and Officials' Attitudes toward Reforms. (a) and (c) present the average psychological measure scores with standard errors, using an ordinal variable where -1 signifies opposition to the reform, 0 indicates a neutral or no explicit attitude, and 1 denotes support for the reform (\textit{N} = 108). (b) and (d) depict the linear regression lines accompanied by 95\% Confidence Intervals, employing a continuous variable that ranges from 0 to 1 to quantify officials' degree of support for the reform (\textit{N} = 56).}
  \label{fig_officials}
\end{figure}

\begin{table}[!htbp]
\centering
\begin{tblr}{
  width = \linewidth,
  colspec = {X[2.5, c]X[2, c]X[3, c]}, %
  hlines, %
}
 & Support for Reform & Attitude toward Reform\\
Traditionalism & -0.441*** & -0.279**\\
Authority & -0.472*** & -0.310**\\
\end{tblr}
\caption{Spearman correlation between CCR-based measure of moral values and actual attitude toward reform of officials. **\textit{p} < .01 ***\textit{p} < .001}
\label{table_officials}
\end{table}

We find a significant correlation (Figure \ref{fig_officials}) between officials' attitudes toward the reforms and the levels of traditionalism and authority measured through CCR.
Authority and traditionalism both show a significant negative correlation with support for reform, with Spearman correlation coefficients less than 0.4 and p-values less than 0.001 (Table \ref{table_officials}). Officials with greater traditionalism and respect for existing authority are more likely to oppose reform, which is in line with the theoretical assumptions.
This benchmarking against historically verified data supports the validity of CCR as a valid computational pipeline to extract meaningful psychological information from classical Chinese corpora.

\section{Discussion and Conclusion}
Historical-psychological text analysis is a new line of research focused on extracting different aspects of psychology from historical corpora using state-of-the-art computational methods \citep{Atari_2023}. Here, we create a new pipeline, CCR, as a helpful tool for historical-psychological text analysis. Evaluating our model against word embedding models (e.g., DDR) and more recent LLMs (e.g., GPT4), we demonstrate that CCR performs better than these alternatives while keeping its high level of interpretability and flexibility. Classical Chinese is of great historical significance, and the proposed approach can be particularly helpful in testing new insights about the ``dead minds'' who lived centuries or even millennia prior. We hope our tool motivates future work at the intersections of psychology, quantitative history, and NLP. 
Importantly, benchmarking historical-psychological tools, especially in ancient languages, is difficult because obtaining ground truth is challenging and dependent upon the quality of historical data. That said, we validate CCR against a historically verified knowledge base about attitudes toward reform and traditionalism.

\section*{Limitation}
Due to the lack of fine-grained data available for training in the context of classical Chinese and with historical-psychological texts, we propose an indirect supervised learning approach where the similarities between titles are used as the pseudo ground truth for similarities between paragraphs. However, this approach may lead to the model learning some noise from the data, negatively affecting the model's performance in downstream tasks.

Our experiments show that hard sampling is counterintuitively worse than random sampling on our dataset (Figure \ref{sampling_compare}). 
This is the case because although the title of a text represents the main idea of most of the content, there may still be parts of the text that are unrelated to the title. For example, in a pair of paragraphs that are identified as positive samples due to their highly similar titles, one paragraph might be irrelevant to the title. Consequently, the text similarity calculated after embedding by a pre-trained model might not be high for this pair of paragraphs. The difference between the similarity prediction made by the pre-trained model and the pseudo ground truth based on title similarity may result in these paragraph pairs being identified as hard samples. However, in such cases, the pre-trained model's prediction could be more accurate than the pseudo ground truth derived from title similarity.
It is the noise caused by the indirect supervised approach that makes the hard sampling fail to find helpful instances.

Our future efforts will be directed toward assembling datasets with expert annotations to address this issue. Moreover, we aim to contribute to both historical psychology and NLP by compiling new open-source datasets for benchmarking purposes.

\bibliography{anthology,custom}
\bibliographystyle{acl_natbib}

\appendix
\clearpage
\newpage

\section{Historical Psychology Corpus Details}
\label{sec:appendix_corpus}

\subsection{Distribution of Paragraph Lengths}
\label{sec:appendix_corpus_para}

To ensure the inclusion of sufficient semantic information, paragraphs containing fewer than 50 characters have been merged with the preceding paragraph of the article or chapter, wherever possible. To accommodate the token limitations of models such as BERT, paragraphs that exceed 500 characters have been divided into segments with fewer than 500 characters each, while maintaining the integrity of the original sentence structure as much as possible. The average length of paragraphs is 195 characters.

\begin{figure}[htbp]
  \centering
  \includegraphics[width=\linewidth]{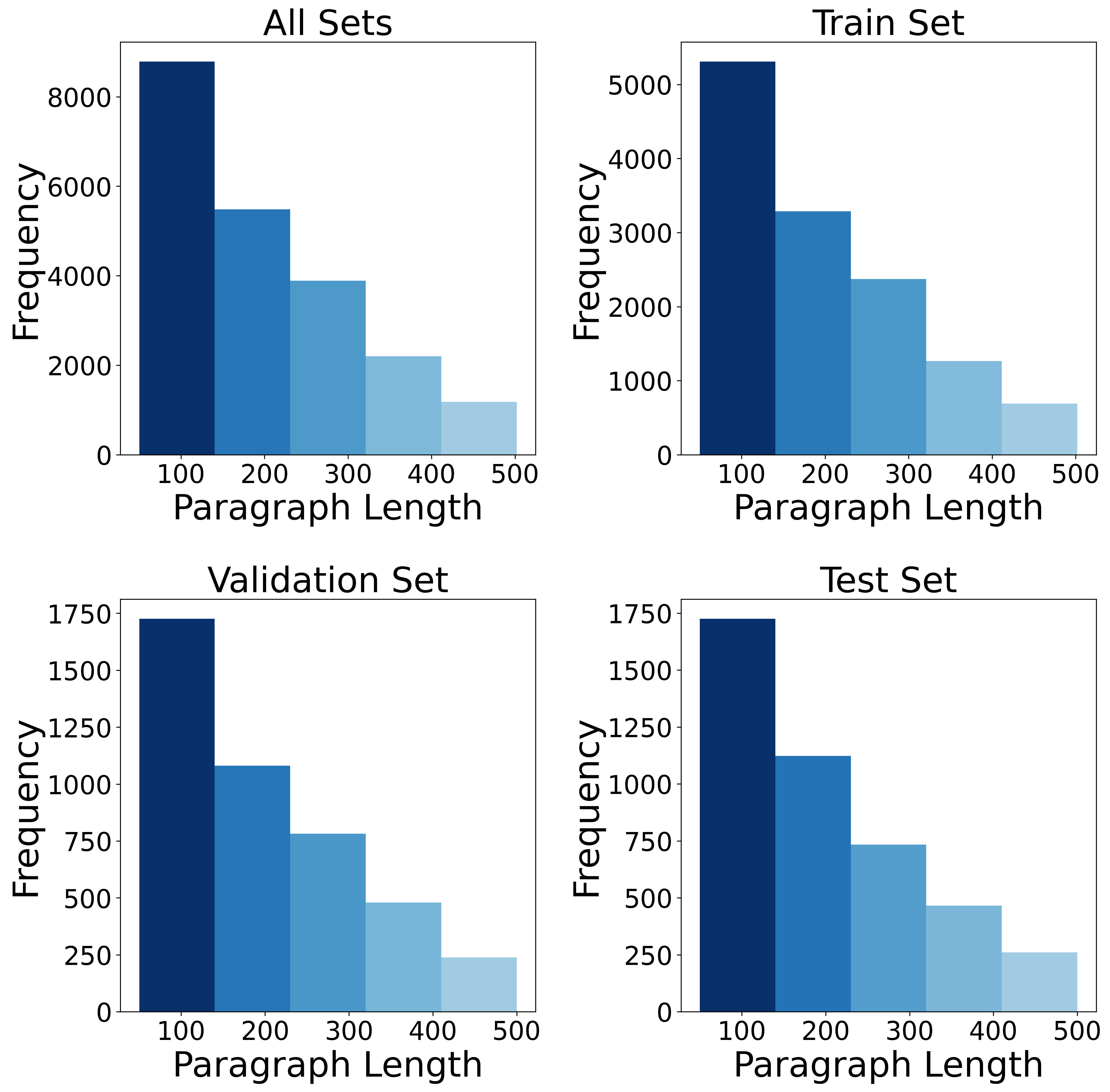}
  \caption{Distributions of paragraph lengths in different sets.}
  \label{para_length_distribution}
\end{figure}

\subsection{Distribution of Title Similarities}
\label{sec:appendix_corpus_title}

\begin{figure}[htbp]
  \centering
  \includegraphics[width=\linewidth]{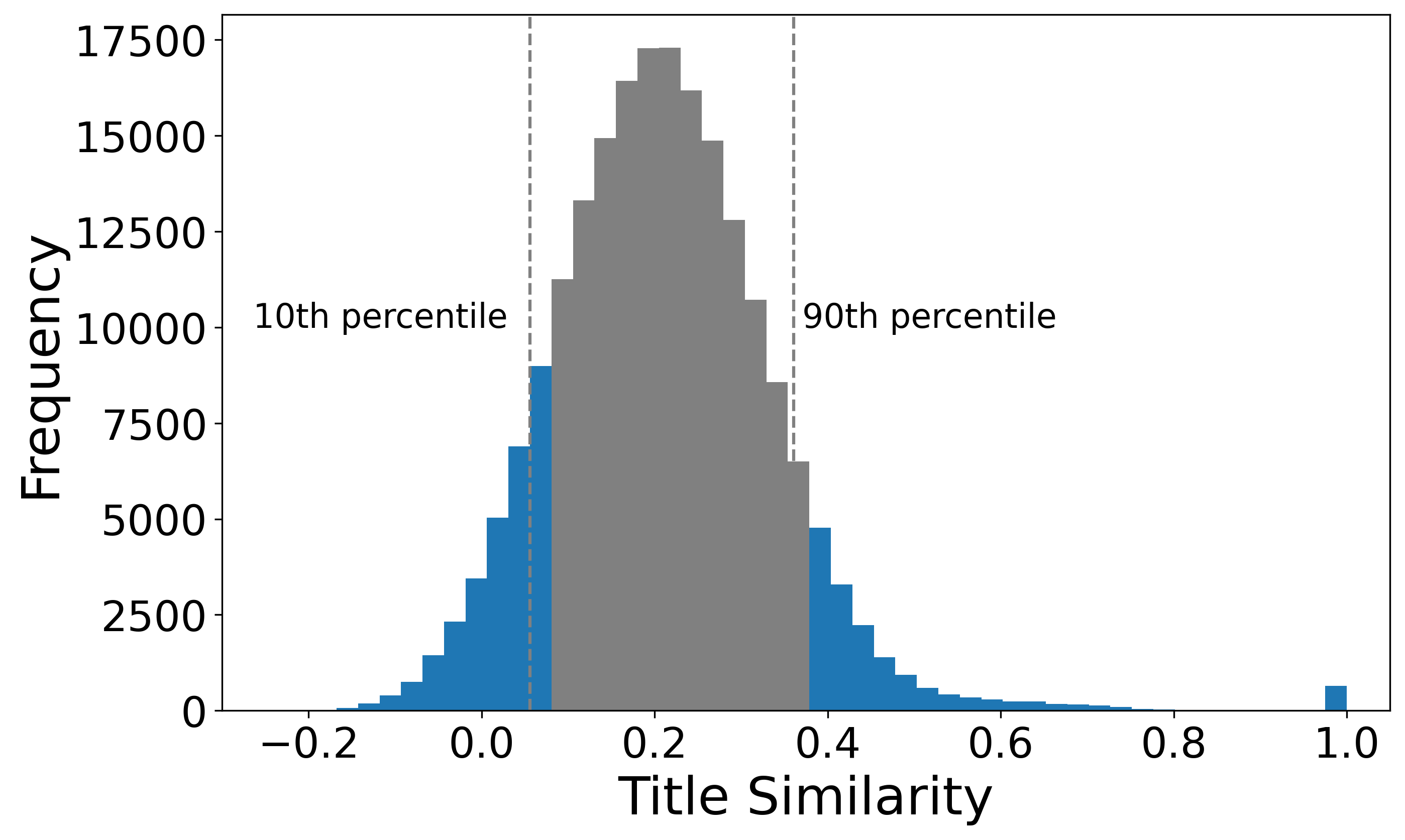}
  \caption{Distribution of title similarities with thresholds.}
  \label{title_similarity_distribution}
\end{figure}

\section{Word Embedding Model Details}
\label{sec:appendix_word_vector_model}

\subsection{Pre-processing}
\label{sec:appendix_word_vector_preprocessing}
Before training the word vector model, we conducted word segmentation on the corpus, employing the pre-trained tokenizer ``\texttt{\path{COARSE_ELECTRA_SMALL_ZH}}'' from HanLP (\url{https://hanlp.hankcs.com/docs/api/hanlp/pretrained/tok.html}).

After word segmentation, the corpus consists of 1.04 billion word tokens and an initial vocabulary containing 15.55 million unique words. By truncating the vocabulary at a minimum word count threshold of 10, the final vocabulary size is reduced to 1.27 million words.

\subsection{Training Hyperparameters}

We train our word vector models on the same corpus using various frameworks and architectures, such as Word2Vec (with CBOW and Skip-gram) \citep{mikolov2013word2vec}, FastText (with CBOW and Skip-gram) \citep{bojanowski2017fasttext}, and GloVe \citep{pennington2014glove}. The hyperparameters are presented in Table \ref{word_vector_models}.

\begin{table*}[!ht]
\centering
\resizebox{\textwidth}{!}{
\begin{tabular}{cccccc}
\toprule[1.2pt]  %
Framework & Architecture & Vector Size & Epoch & Window Size & Other Parameters \\ \hline
\multirow{2}{*}{Word2Vec} & CBOW & 300 & 5 & 5 & negative=5 \\ \cline{2-6} 
 & Skip-gram & 300 & 5 & 5 & negative=5 \\ \hline
\multirow{2}{*}{FastText} & CBOW & 300 & 5 & 5 & negative=5, min\_n=1, max\_n=4 \\ \cline{2-6} 
 & Skip-gram & 300 & 5 & 5 & negative=5, min\_n=1, max\_n=4 \\ \hline
GloVe &  & 300 & 15 & 5 & x\_max=100, alpha=0.75 \\ 
\bottomrule[1.2pt]  %
\end{tabular}
}
\caption{Word vector model training hyperparameters and evaluation results.}
\label{word_vector_models}
\end{table*}

\section{Dictionary Details}
\label{sec:appendix_dictionary}

We build a dictionary for each classical Chinese questionnaire by using an open-source dictionary system named ``WantWords'' \cite{qi2020wantwords}, which is based on a multi-channel reverse dictionary model (MRDM) \cite{zhang2020multi} and takes sentences (descriptions of words) as input and yields words semantically matching the input sentences. 

The process involves three steps: 
(1) we employ the ``WantWords'' model to obtain the top n most similar words to each quotation in the questionnaire;
(2) a process of deduplication is then conducted;
(3) the words are labeled manually by a native Chinese speaker with ``relevant'' or ``irrelevant'' to the corresponding topic, after which all irrelevant words are discarded.

\begin{table*}
\centering
\begin{tblr}{
  colspec = {X[0.48,c,m]X[0.92,c,m]X[1,c,m]},
  width = \linewidth,
  hlines,
  hline{1} = {1.2pt},
  hline{Z} = {1.2pt},
  cell{1}{1} = {r=3}{c,m},
  cell{4}{1} = {r=5}{c,m},
}
Sampling & Positive/Negative Sampling Thresholds & \{(10th, 90th)\} \\
         & Triplet Sampling Option & \{random\} \\
         & Sampling Seed & \{42\} \\
Training & Batch Size & \{16, 32\} \\
         & Epochs & \{3\} \\
         & Warmup Epochs & \{1, 2, 3\} \\
         & Learning Rate & \{1e-6, 1e-5, 2e-5\} \\
         & Optimizer & \{Adam\} \\
\end{tblr}
\caption{Hyperparameter sweep for triplet sampling and validation for fine-tuned models.}
\label{hyperparameter_sweep}
\end{table*}

\clearpage

\begin{figure}[htbp]
  \centering
  \includegraphics[width=\linewidth]{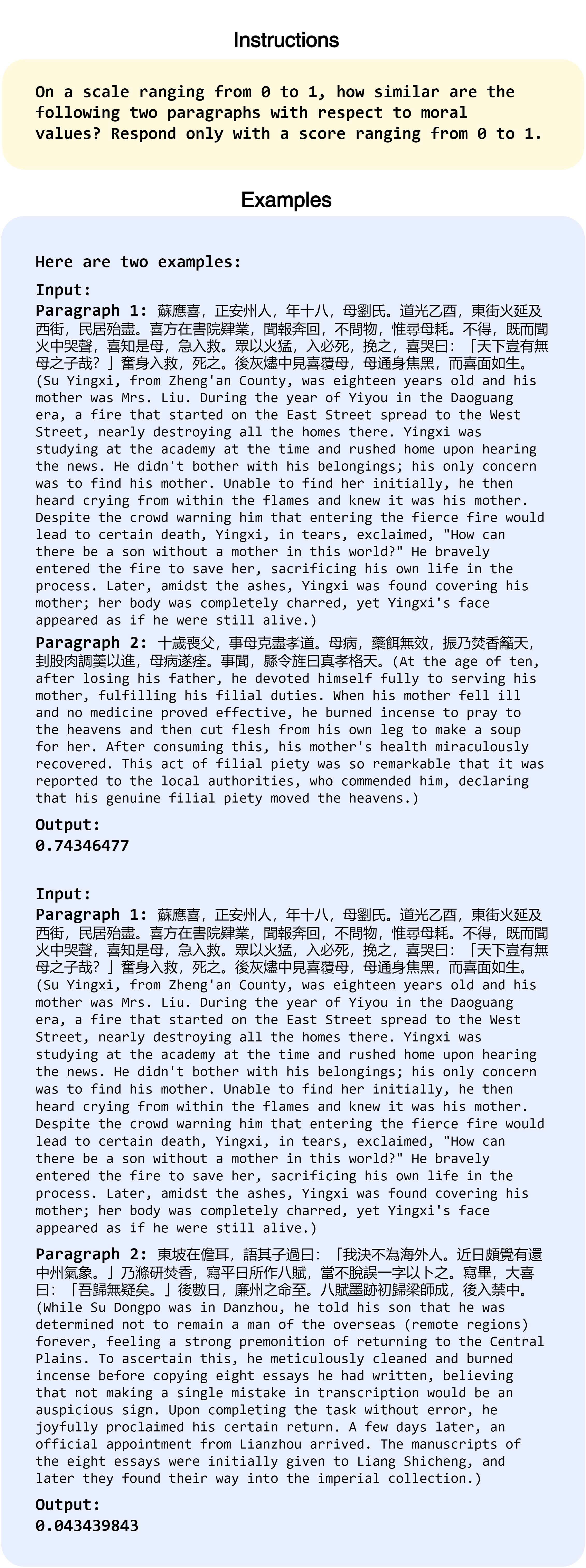}
  \caption{Few-shot prompt for the semantic textual similarity task.}
  \label{prompt_similarity}
\end{figure}

\begin{figure}[htbp]
  \centering
  \includegraphics[width=0.9\linewidth]{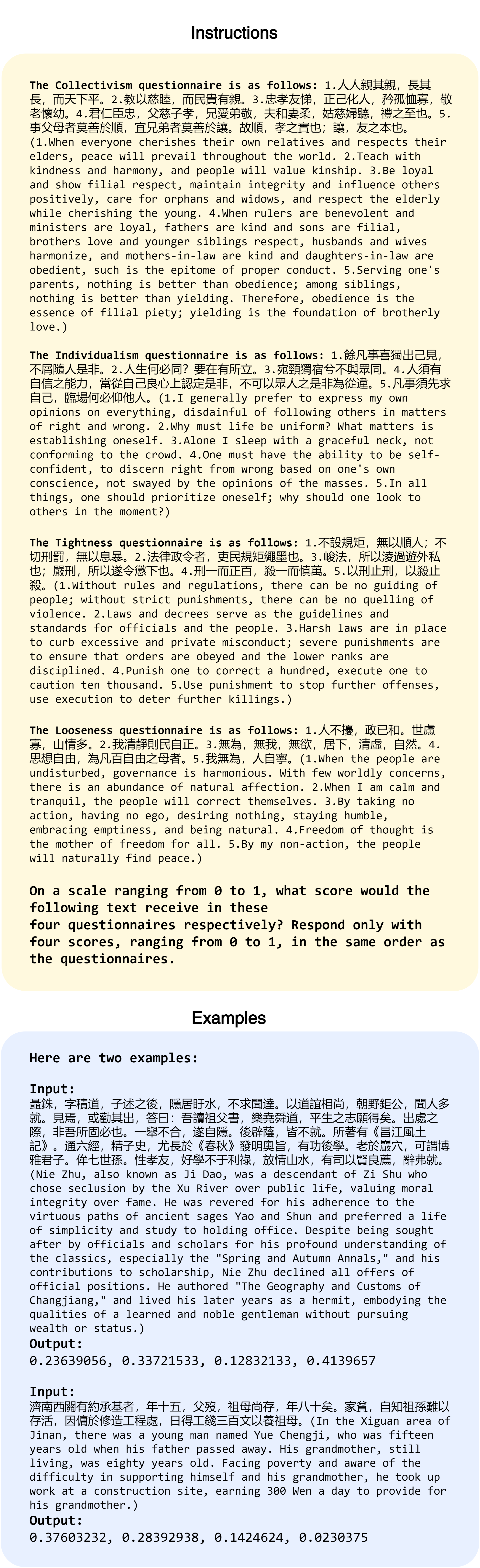}
  \caption{Few-shot prompt for the psychological measure task.}
  \label{prompt_measure}
\end{figure}

\begin{figure*}[htbp]
  \centering
  \includegraphics[width=0.6\linewidth]{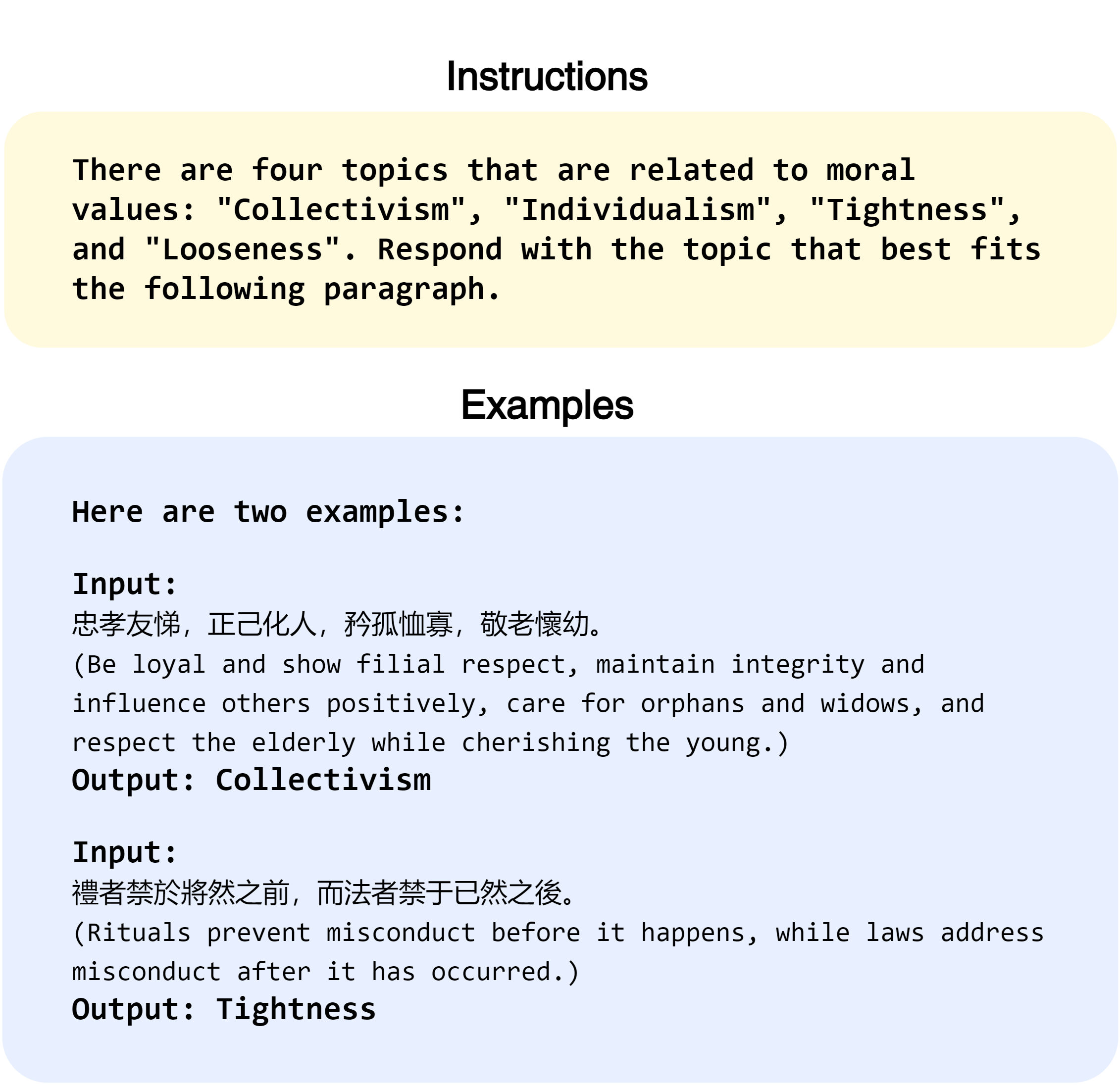}
  \caption{Few-shot prompt for the questionnaire item classification task.}
  \label{prompt_classification}
\end{figure*}

\end{document}